\documentclass[journal]{IEEEtran}

\usepackage[detect-all=true,separate-uncertainty=true, separate-uncertainty]{siunitx}
\usepackage{cite}
\usepackage{hyperref}
\usepackage[pdftex]{graphicx}
\DeclareGraphicsExtensions{.pdf,.jpeg,.png}
\usepackage{amsmath}
\usepackage{algorithmic}
\usepackage{wasysym}
\usepackage[normalem]{ulem}
\usepackage{array}
\usepackage[dvipsnames,table]{xcolor}
\ifCLASSOPTIONcompsoc
  \usepackage[caption=false,font=normalsize,labelfont=sf,textfont=sf]{subfig}
\else
  \usepackage[caption=false,font=footnotesize]{subfig}
\fi
\usepackage{url}
\usepackage{microtype}
\usepackage{enumitem}  
\setlist[description]{font=\normalfont\itshape}  
\usepackage{booktabs}
\usepackage{menukeys}
\usepackage{marginnote}

\newcommand{\figref}[1]{Fig.~#1}
\newcommand{\tabref}[1]{Table~#1}
\newcommand{\secref}[1]{Section~#1}
\newcommand{\linktocode}{\href{https://github.com/StefanTUHH/robotic_needle_insertion}{www.github.com/StefanTUHH/robotic\_needle\_insertion}}

\usetikzlibrary{arrows,positioning,shapes.geometric}
\usetikzlibrary{intersections}
\definecolor{customblue}{RGB}{0,0,128}

\begin{document}

\title{Robotic Tissue Sampling for Safe Post-mortem Biopsy in Infectious Corpses}
\author{Maximilian~Neidhardt*,
        Stefan~Gerlach*,
        Robin~Mieling*,
        Max-Heinrich~Laves,
        Thorben~Wei\ss{},\\
        Martin~Gromniak,
        Antonia~Fitzek, 
        Dustin~M{\"o}bius, 
        Inga~Kniep, 
        Alexandra~Ron, 
        Julia~Sch{\"a}dler,\\
        Axel~Heinemann,
        Klaus P\"uschel,
        Benjamin~Ondruschka,
        and~Alexander~Schlaefer
\thanks{\textbf{Maximilian~Neidhardt, Stefan~Gerlach, Robin~Mieling, Max-Heinrich~Laves, Martin~Gromniak and Alexander~Schlaefer,}  Institute of Medical Technology and Intelligent Systems, Hamburg University of Technology, Germany (e-mail: Maximilian.Neidhardt@tuhh.de)}%
\thanks{\textbf{Thorben~Weiß, Antonia~Fitzek, 
        Dustin~M\"obius, Inga~Kniep, Alexandra~Ron, Julia~Sch\"adler, Axel~Heinemann, Klaus P\"uschel and  Benjamin~Ondruschka,} Institute of Legal Medicine, University Medical Center Hamburg-Eppendorf}
\thanks{Manuscript received August 31, 2021; revised January 01, 2022.}
\thanks{* Authors contributed equally.}}

\markboth{Journal of IEEE Transactions on Medical Robotics and Bionics,~Vol.~XX, No.~X, August~2021}%
{Journal of IEEE Transactions on Medical Robotics and Bionics,~Vol.~XX, No.~X, August~2021}

\IEEEspecialpapernotice{Special Issue on Surgical Vision, Navigation, and Robotics}

\maketitle

\begin{abstract}
In pathology and legal medicine, the histopathological and microbiological analysis of tissue samples from infected deceased is a valuable information for developing treatment strategies during a pandemic such as COVID-19. However, a conventional autopsy carries the risk of disease transmission and may be rejected by relatives. We propose minimally invasive biopsy with robot assistance under CT guidance to minimize the risk of disease transmission during tissue sampling and to improve accuracy. A flexible robotic system for biopsy sampling is presented, which is applied to human corpses placed inside protective body bags. An automatic planning and decision system estimates optimal insertion point. Heat maps projected onto the segmented skin visualize the distance and angle of insertions and estimate the minimum cost of a puncture while avoiding bone collisions. Further, we test multiple insertion paths concerning feasibility and collisions. A custom end effector is designed for inserting needles and extracting tissue samples under robotic guidance. Our robotic post-mortem biopsy (RPMB) system is evaluated in a study during the COVID-19 pandemic on 20 corpses and 10 tissue targets, 5 of them being infected with \linebreak SARS-CoV-2. The mean planning time including robot path planning is \SI{5.72(167)}{\second}. Mean needle placement accuracy is \SI{7.19(422)}{\milli\meter}.
\end{abstract}

\begin{IEEEkeywords}
Collaborative Robot, Path Planning, COVID-19, Forensic Medicine,Medical Robotics
\end{IEEEkeywords}

\section{Introduction}

\IEEEPARstart{I}{n} June 2021, the World Health Organization reported that the number of deaths from COVID-19 has surpassed 4 million people. Nonetheless, vaccines have shown to be effective in defeating severe courses and fatal outcomes of the pandemic. However, the treatment of serious infections remains complex due to the multi-faceted nature of the pathophysiology of COVID-19 with various disease stages \cite{vanEijk.2021}. Studying pathological changes of multiple organs from deceased is crucial for understanding the disease progression and to develop new treatments \cite{Aghayev.2007}. Hence, multiple studies examining tissue biopsies of COVID-19 deceased have been performed with a focus on cardiovascular \cite{Wenzel.2020,Lindner.2020} and lung tissue \cite{Heinrich.2021} as well as multi-organ dysfunctions \cite{DeinhardtEmmer.2021}.

\begin{figure}[t]
    \centering
    \includegraphics[width=\linewidth]{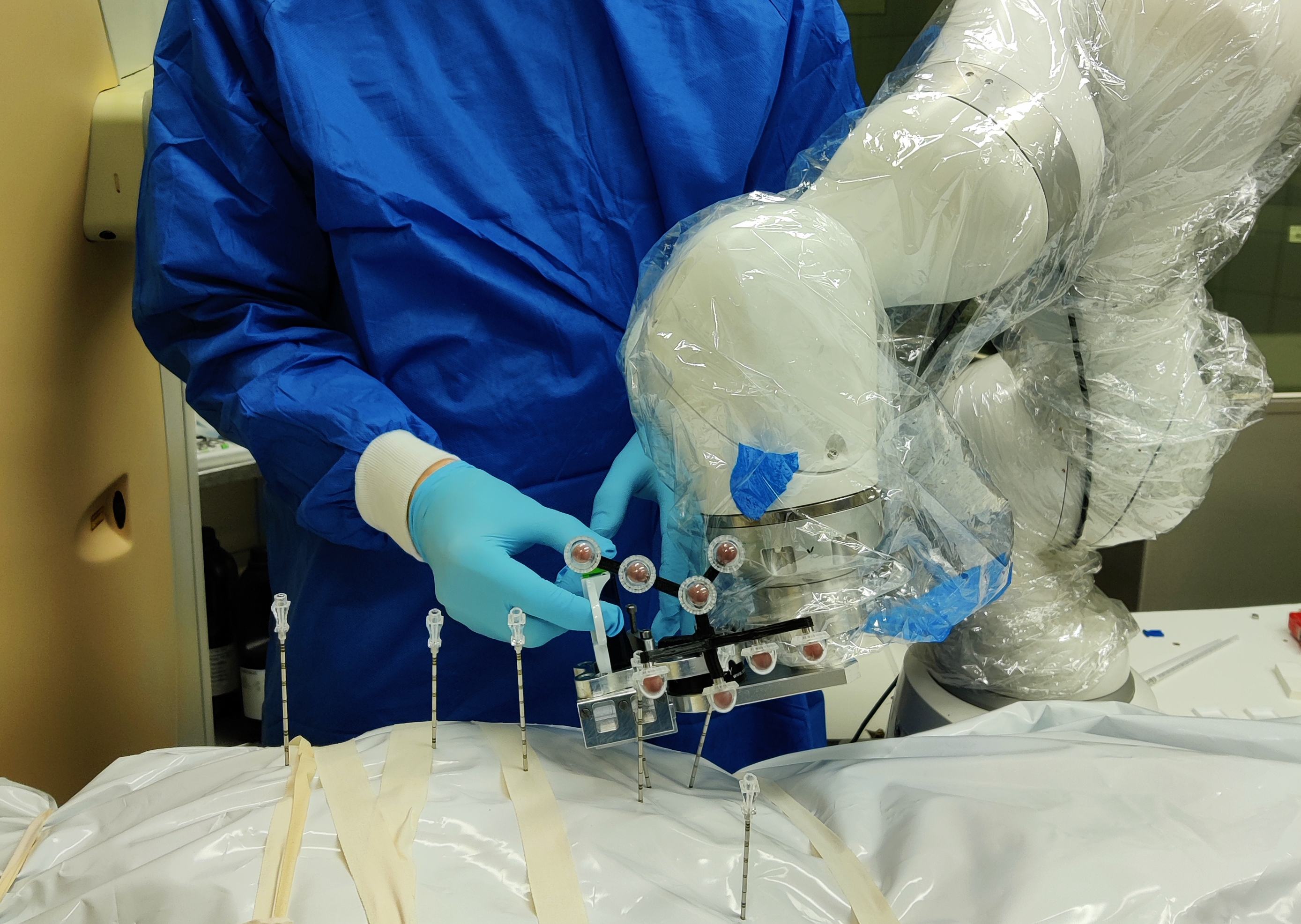}
    \caption{\textbf{Minimally invasive tissue sampling with a robot.} The robot drives a biopsy needle into the corpse to the desired target. The physician can then take a biopsy sample while the robot holds the needle with our specially designed needle holder. A protective body bag reduces the risk of disease transmission. }
    \label{fig:System_Setup_1}
\end{figure}

Conventionally, tissue samples for histological, immunochemical and microbiological investigations are collected from deceased during an autopsy. However, recent studies have shown that personal protective equipment is contaminated during conventional autopsies with the SARS-CoV-2 virus, showing replicability in some cases and therefore a potential risk \cite{Brandner.2021}. Recommendations from the German National Infection Surveillance authorities (Robert- Koch Institute) had been hesitant regarding performance of autopsies at the onset of the COVID-19 pandemic. Moreover, conventional biopsies are time consuming and consent from relatives might be denied, e.g., for religious reasons or in case of children \cite{Weustink.2009}. Minimally invasive techniques with biopsy needles employed for probing of tissue or fluids through small punctures is a much less destructive approach. 

\begin{figure*}[t]
    \centering
    \includegraphics[width=\linewidth]{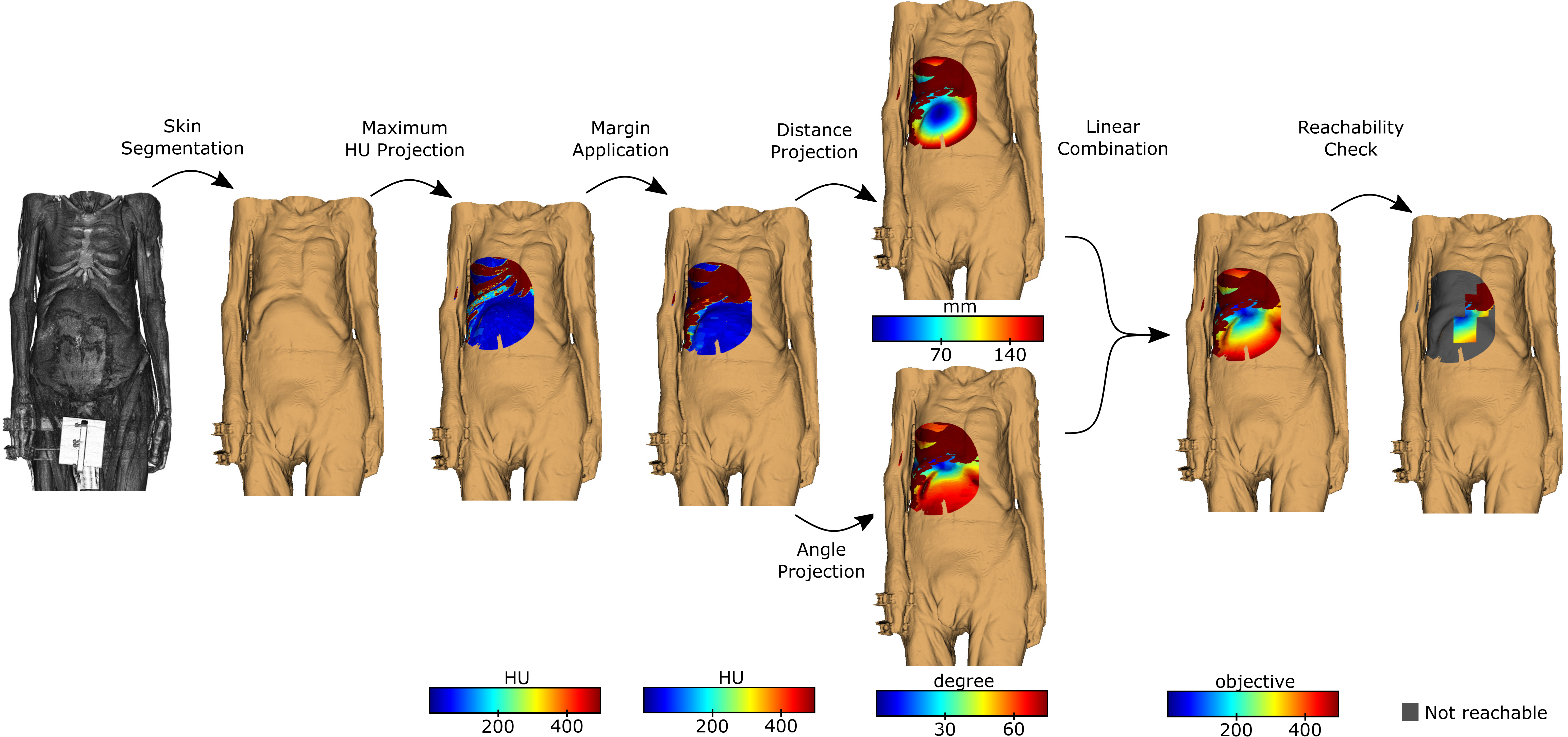}
    \caption{\textbf{Steps to visualize feasible robot insertion positions.} Illustration of iterative overlay generation. Every arrow represents an operation applied to the previous overlay. Skin is segmented, maximum CT density is estimated, margin is added, distance and insertion angle are estimated, and reachability is verified. Skin colored points are outside of needle range, dark red points are occluded by dense tissue, grey points cannot be reached, and blue points have low objective value.}
    \label{fig:pipeline_Overlay}
\end{figure*}

In clinical routine, medical needles are often placed by hand under ultrasound, CT or MRI guidance, e.g., for drug delivery or conventional biopsy. As a consequence of the COVID-19 pandemic, minimally invasive biopsy under US guidance was also considered for autopsies to reduce the risk of disease spreading. However, the extraction of tissue probes is strongly operator-dependent and a direct contact between physician and the infectious cadaver cannot be avoided, leaving the risk of possible disease transmission. Moreover, conventional methods are time consuming for medical experts, especially when a large number of tissue sets are sampled for biobanking to enable systematic and multidisciplinary research.

Robots are a promising alternative for minimally invasive autopsies to enhance user independent needle placement accuracy and perform iterative and safe biopsy sampling \cite{Siepel.2021}. Many systems for medical needle insertion can be found in the literature which pre-align the needle while the insertion is performed manually by the physician \cite{Kettenbach.2014, Levy.2021, Martinez.2014}. Other systems execute manually planned insertions in phantoms, animal models, or patients \cite{BenDavid.2018,Hiraki.2017,Hiraki.2020} for specific biopsy scenarios. Still, biopsy sampling in a post-mortem setup is different, since needle insertions through, e.g., major blood vessels are feasible. Moreover, time efficiency during biopsy extraction is more essential to reduce the case load for medical staff and much larger sample sets are collected than in the living. Overall, robotic systems are ideal for needle placement, but a decision support software is necessary to allow fast, safe and simple trajectory planning and interaction with the robot.

In this work, we present the \textbf{ro}botic \textbf{p}ost-\textbf{m}ortem \textbf{b}iopsy (RPMB) system which enables safe tissue extraction with an off-the-shelf 7 degrees-of-freedom (DOF) light-weight robot (LBR Med 14, KUKA). The additional DOF enables biopsy for a wide range of targets with limited space. In the RPMB system, the robot drives a biopsy needle into a corpse which is wrapped in a protective body bag. Thereby, we limit infection risks of staff members and the time being in close contact to the corpse. For guidance, we use a post-mortem computed tomography (PMCT) taken prior to the biopsy. We implement our approach as a module integrated in an open source software for medical image processing (3D Slicer, \cite{Fedorov.2012}).

The goal of this study is (I) to present a real-world post-mortem clinical setup for minimal invasive robotic autopsies with an intuitive to use planning software and (II) to demonstrate the clinical workflow for extracting a set of tissue biopsies from corpses with the RPMB system.

For our first goal, we present and evaluate a path planning module to find the optimal insertion point with a robot. Feasible needle paths are identified by considering the needle length, avoiding bone contact during insertion, and verifying robot path feasibility without a collision with the corpse, the robot, or the CT gantry and patient table. Several clinical decision support systems are described in the literature, which assist the physician in placing needles during very specific interventions, e.g., resections, biopsy extractions or brain surgery \cite{LeonCuevas.2017,Beriault.2011,Baegert.2007,Villard.2005}. In contrast to clinical interventions, extracting biopsies in corpses permits a direct access to the target while avoiding bone punctures, since needle punctures of delicate structures, such as blood vessels or nerves, are of no importance. Therefore, avoiding bone punctures, minimizing procedure times, and sufficient needle placement accuracy are desirable but have not been the focus of previous work. Ebert et al.\ have demonstrated a decision support system for their ceiling-mounted robotic system ``Virtobot'' \cite{Ebert.2010}. Here the authors focused on a small number of targets with a system requiring significant adjustments to the operation hall, i.e., a permanently installed ceiling mounted system. In contrast, the light-weight robot mounted on a movable table is highly flexible. The RPMB system can be removed completely from the examination room to not intervene with other medical procedures while the setup including a one-time calibration can still be done in under approximately 30 minutes.

\begin{figure*}
    \centering
    \includegraphics[width=.99\linewidth]{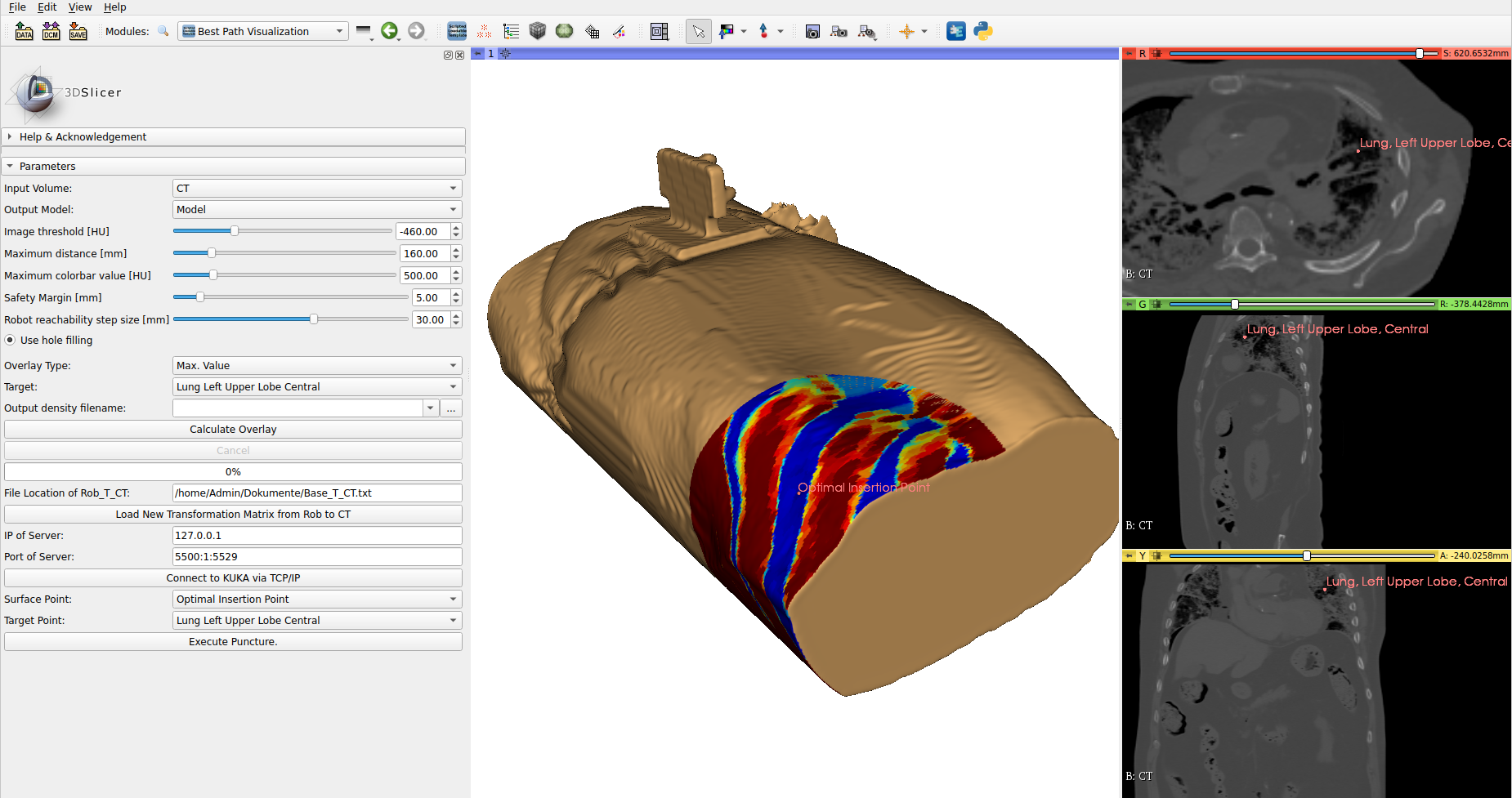}
    \caption{\textbf{Best path visualization module in 3D Slicer.} The module segments the skin of the corpse and projects colormaps for various overlay types, e.g., bone density or distance to target onto the skin. The colors in this example indicate the maximum CT values along the needle insertion path. The module and its code can be downloaded from: \linktocode.}
    \label{fig:slicer_screenshot}
\end{figure*}

For our second goal, we demonstrate and evaluate a real-world RPMB system setup for automatic needle insertion and show the extraction of biopsy samples. The optimal needle path is determined from a set of feasible and reachable paths and a workflow is presented which can be transferred to other specialized centres for post-mortem imaging. We demonstrate, that with the RPMB system we can acquire a set of tissue biopsies for 10 different organs from 20 corpses with an average accuracy of \SI{7.19(422)}{\milli\meter} without direct contact to the corpses.

\section{Methods}
The methods section is structured as follows: First, we present a method for estimating feasible insertion points and a graphical user interface (GUI) for visualization and insertion planning. Second, we present a clinical workflow for post-mortem biopsy sampling and third, we describe the evaluation of the RPMB system with respect to target accuracy and biopsy quality in an extensive clinical study on corpses.

\subsection{Planning and Decision System}
For path planning, we first identify suitable insertion points on the skin of the corpse and validate which positions are reachable by the robot.

\begin{figure*}[t]
    \centering
    \includegraphics[width=\linewidth]{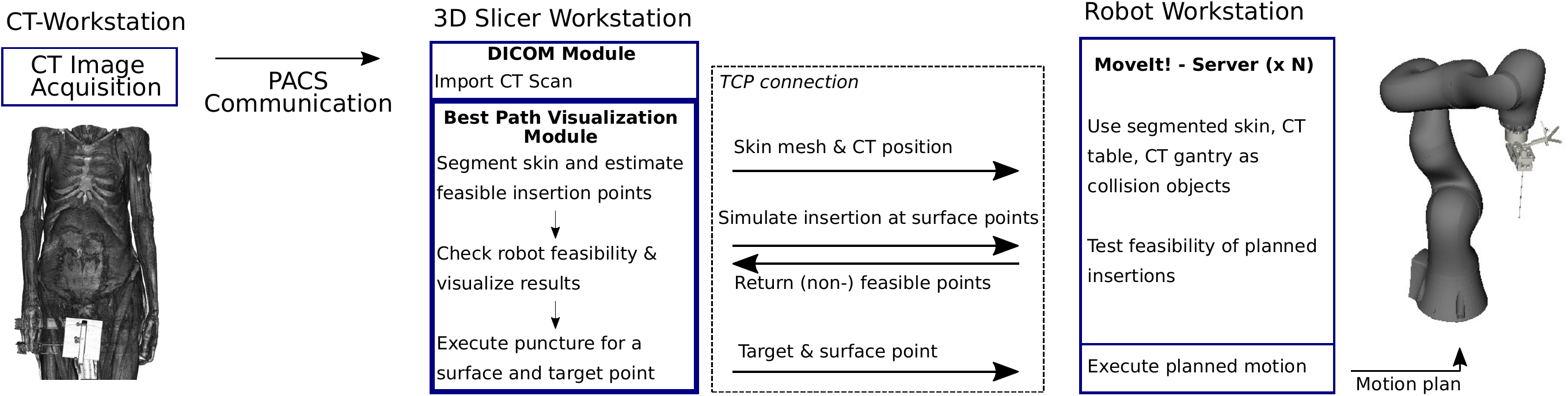}
    \caption{\textbf{Communication between workstations.} Overview of communication and module setup. Three different application bundles are running (CT acquisition, 3D Slicer, MoveIt ecosystem. Note that applications can run on separate computers and multiple (N) instances of the MoveIt ecosystem can be executed to parallelize reachability evaluation---even across multiple computers.}
    \label{fig:System_Communication}
\end{figure*}

\subsubsection{Insertion Path Visualization}
\label{sec:Insertion_Path_Visualization}
To identify feasible entry points and provide an intuitive presentation, we generate an overlay onto the skin of the corpse showing the feasibility and quality of entry points. The individual steps, for which an example is shown in \figref{\ref{fig:pipeline_Overlay}}, are the following:
\begin{description}
    \item[Skin Segmentation:] The skin is segmented using thresholding followed by morphological operations to refine the segmentation. First, the largest connected volume obtained by a threshold based segmentation is identified. After a closing operation with kernel size of \SI{1}{cm} on the segmented volume to close openings, e.g., the nostril, we invert the segmentation. Thereby, only non-tissue is segmented. We again identify the largest continuous segmentation (air outside of the corpse) and invert the segmentation resulting in closing the previously unsegmented volume below the threshold inside the corpse, e.g, inside the lung. Finally, we generate the surface of the segmentation.
    \item[Maximum HU projection:] The set of feasible insertion points on the segmented skin is restricted to points with distances to the target smaller than the length of the needle $l$. We also remove insertion points if additional body parts are located on the direct path in air towards the target, e.g., due to obstruction by an arm. For the remaining points we calculate the maximum Hounsfield units (HU) along the line from the target to the skin surface.
    \item[Margin Application:] To avoid punctures of the bone, we apply an additional safety margin of \SI{5}{\milli\meter} around projected dense structures, e.g., bones or further needle-resistent objects (medical implants, foreign bodies) by applying a dilation filter.
    \item[Distance and Angle Projection:] We linearly combine the insertion depth $d$ and the insertion angle to the skin's normal $a$ for feasible insertion points as a cost measure $q = 0.5 \cdot \hat{d} \; + \; 0.5 \cdot \hat{a}$, where variables are normalized by $\hat{d} = d \cdot \frac{t}{l}$ and $\hat{a} = a \cdot \frac{t}{90}$. We define the optimum point for an insertion as the point with the lowest cost. The goal is to avoid large needle path distances and sharp angles on the skin, which can reduce needle placement accuracy.
    \item[Reachability Check:] feasible insertion points are checked for reachability by the robot. To reduce computational effort, we sample feasible points on a grid along the CT axes and simulate the needle insertion procedure for each grid point. We evaluate grid points in parallel to decrease computation time.
\end{description}

\subsubsection{Implementation as 3D Slicer Module}
We implement the overlay generation described in \secref\ref{sec:Insertion_Path_Visualization} as a 3D Slicer module. 3D Slicer \cite{Fedorov.2012} is an open-source development and research platform for medical image processing tasks. It enables development and distribution of application specific solutions. Our code for the module developed in this project can be found at \linktocode. 

\figref{\ref{fig:slicer_screenshot}} shows the graphical user interface in 3D Slicer. Here, the user can load a PMCT from the database and adjust relevant parameters, e.g., the skin segmentation threshold, the maximum length of the biopsy needle or the additional safety margin for dense structures. Moreover, several types of overlays are available which are explained in detail in our previous work \cite{BMT_NEIDHARDT_GERLACH}. The transformation matrix between the robot base and the CT reference system can be loaded if available. The overlay heat map is calculated and a 3D rendering of the skin model with possible insertion points is shown, similar to \figref\ref{fig:slicer_screenshot}, right. The point with the minimal cost for the selected overlay type is computed and indicated on the skin model. The physician can now, if necessary, relocate the insertion point either in the conventional CT image slices or directly on the skin rendering. For executing the puncture, the insertion and target point are sent to an external robot control server.

\subsubsection{Robot Path Planning}
\label{sec:Robot_Path_Planning}
\begin{figure*}[bt]
    \centering
    \includegraphics[width=\linewidth]{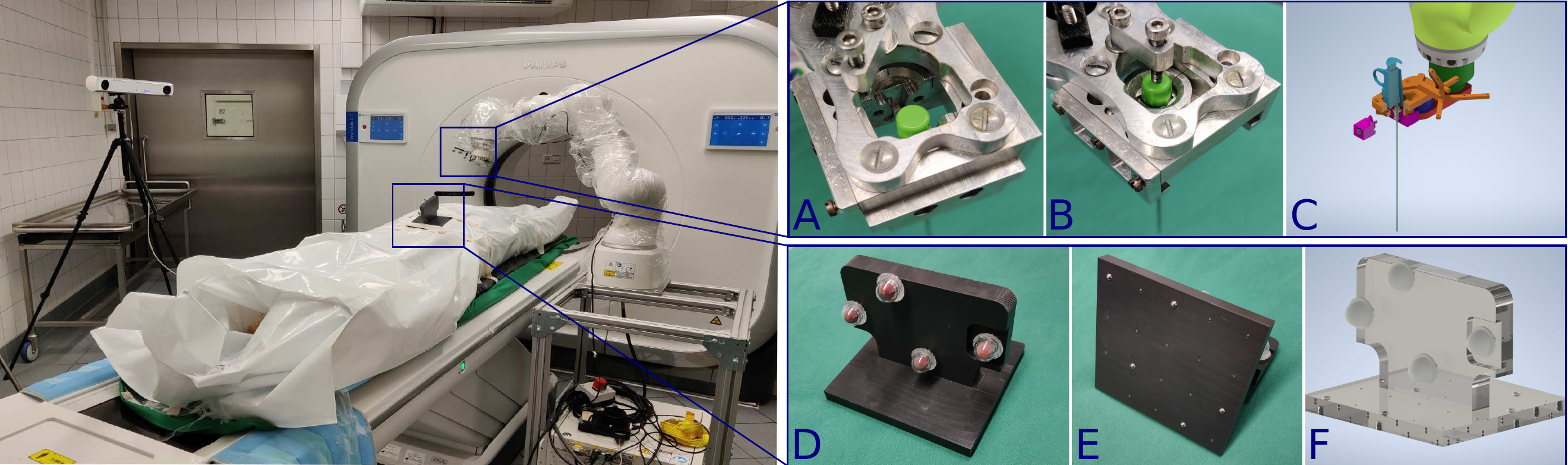}
    \caption{\textbf{RPMB System Setup.} \textit{Left:} To prevent disease contamination, we place infectious corpses in body bags. A robot (LBR Med 14, KUKA) inserts biopsy needles into the corpse through the body bag. \textit{Top right:} [A] Custom robot end effector for mounting a hollow guide needle in which we insert a co-axial needle. [B] A screw prevents slipping along the needle axis of the co-axial needle during insertions and two additional screws (not visible) on the side tighten the clamping jaws. [C] CAD model with reflective tracking markers (orange) and inserted biopsy needle. \textit{Bottom right:} Registration marker for estimating the position of the CT coordinate system relative to the robot's world coordinate system. The marker orientation can be tracked with [D] the tracking camera and [E] in the CT imaging system with a steel ball checkerboard, [F] the inner transform can be estimated from the CAD model.} 
    \label{fig:System_Setup_2}
\end{figure*}

The RPMB system involves three workstations: (1) clinical CT workstation for image acquisition, (2) 3D Slicer workstation for planning needle insertions in the 3D CT scan, and (3) the robot workstation for checking feasibility and executing needle insertions. Therefore, we designed a robot control server and employ MoveIt! \cite{ColemanDavid.2014} for robot control. MoveIt! is a framework for robotic applications and includes implementations of generic forward/inverse kinematics solvers and methods for path and trajectory planning and execution with support for collision avoidance. Through MoveIt! we also make use of third-party libraries including the Kinematics and Dynamics Library (KDL) \cite{H.Bruyninckx.2012}, the Fast Collision Check Library (FCL) \cite{J.Pan.2012}, and the Open Motion Planning Library (OMPL) \cite{Sucan.2012}. We define the segmented skin of the corpse and a simple plane for the CT gantry as a collision object in MoveIt!. Moreover, we add the vector-graphics model of our custom end effector with a needle, depicted in \figref{\ref{fig:System_Setup_2}}, as an additional link on the robot to enable collision avoidance. We integrate MoveIt! in a separate server structure to check for feasible insertion paths, setting collision objects, and executing insertions. CT data is transferred via the picture archiving and communication system (PACS) and robot communication is realized with the Robot Operating System (ROS) \cite{StanfordArtificialIntelligenceLaboratoryetal..}. The overall information flow for performing a needle insertion is show in \figref\ref{fig:System_Communication} and described in the following: 

\begin{enumerate}[leftmargin=11mm,label=\arabic*.]
    \item A CT scan is acquired and transferred to the 3D-Slicer workstation over a connection to the PACS server (3D-Slicer DICOM module).
    \item A desired target point is defined in the conventional three CT planes by a medical expert, as seen in \figref\ref{fig:slicer_screenshot}.
    \item The skin is segmented and feasible insertion points based on the HU values are estimated. The skin mesh and the CT position relative to the robot's base frame are sent via a Transmission Control Protocol (TCP) connection to the robot workstation and defined as collision objects. Note that multiple instances can run on the robot workstation or multiple workstations to parallelize reachability checks.
    \item Reachability of surface points is checked by simulating needle insertion with the target and the set of surface points. A response containing results of reachability checks for each surface point is returned over TCP.
    \item An optimal insertion point according to the objective function is suggested, which can be adapted by the medical expert if desired. To reduce computation time we reduce the CT data size by half through linear interpolation in each dimension. A target and surface point is chosen and sent to the robot workstation.
    \item A motion plan is estimated, feasibility is checked and executed by the robot.
\end{enumerate}

\begin{figure*}[t!]
    \centering
    \begin{tikzpicture}[>=latex']
        \tikzset{block/.style= {draw, color= customblue, text = black, rectangle, font=\scriptsize, align=center,minimum width=1.5cm,minimum height=1.0cm},
        rblock/.style={draw, font=\scriptsize, shape=rectangle,rounded corners=1.5em,align=center,minimum width=1.5cm,minimum height=1.0cm},
        decision/.style={draw, thick, font=\scriptsize, shape=diamond,align=center,minimum width=1.5cm,minimum height=1.0cm},
        }
        \node [block]  (setup) {System Setup \& \\ Calibration};
        \node [block, right =1cm of setup] (position) {Position corpse \\ in supine or\\ prone position};
        \node [block, right =1cm of position] (acquire) {CT \\acquisition};
        \node [rblock, below =0.8cm of acquire] (params) [font=\tiny] {tube voltage: 120 kV \\ tube current: 296 mAs\\slice thickness: 0.67 mm\\spacial res. z-axis: 0.45 mm\\pixel spacing: 0.98 mm };
        \node [block, right =1cm of acquire] (prepare) {Target annotation \&\\Estimate $_{\text{SB}}T^{\text{CT}}$ };
        \path [draw,dashed] (acquire) edge (params);
        \node [block, right =1cm of prepare] (compute) {Compute optimal\\insertion point};
        \node [decision, right =of compute,inner sep=0.0pt] (check1) {Can robot\\reach target?};
        \node [decision, above = 0.5cm of compute,inner sep=1.5pt] (check2) {All targets\\sampled?}; 
        \node [block, below =0.5cm of check1] (move) {Move CT table \&\\Estimate $_{\text{B}}T^{\text{CT}}$};
        \node [block] at (check1 |- check2) (extract) {Extract\\biopsy};
        \path[draw,->,thick] (setup) edge (position)
                    (position) edge (acquire)
                    (acquire) edge (prepare)
                    (prepare) edge (compute)
                    (compute) edge (check1)
                    (extract) edge (check2)
                    ;                    
        \path[draw,->,thick] (check2)  -| node [near start, above, font=\scriptsize] {Yes}(position);            
        \path[draw,->,thick] (move)  -| (compute);            
        \path[draw,->,thick] (check1) -- node [midway,right, font=\scriptsize] {No} (move);               
        \path[draw,->,thick] (check1) -- node [midway,right, font=\scriptsize] {Yes} (extract); 
        \path[draw,->,thick] (check2) -- node [midway,right, font=\scriptsize] {No} (compute);
    \end{tikzpicture}   
    \caption{\textbf{Biopsy workflow.} The workflow for robotic biopsy sampling with the RPMB system. Note that system setup and calibration only has to be done once and can be reused after repositioning a corpse and for consecutive corpses.}
    \label{fig:systemWorkflow}
\end{figure*}
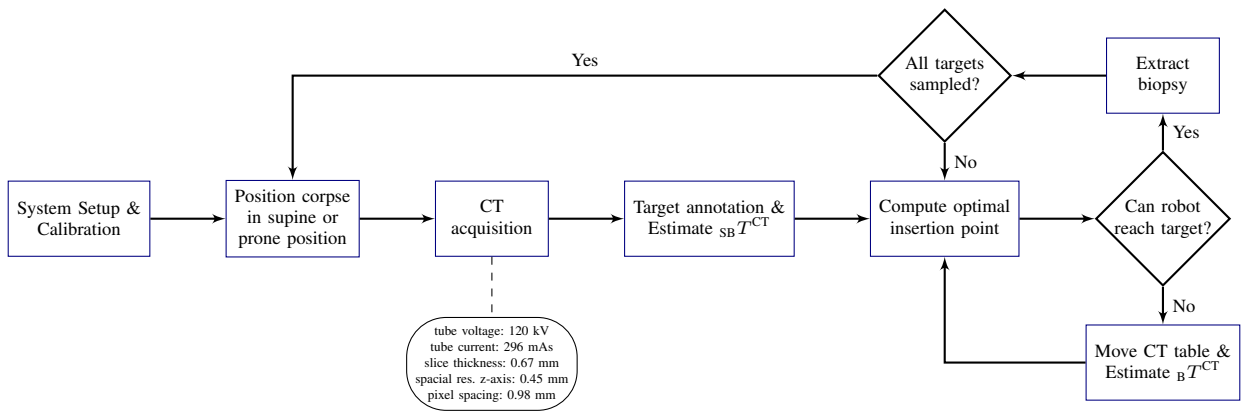

\subsection{Clinical Workflow for Post-Mortem Sampling}
\label{sec:clinical_workflow}
A flow chart of our proposed workflow is shown in \figref{\ref{fig:systemWorkflow}}. The RPMB system is set up and a corpse is positioned on the CT table by the medical staff. A CT scan is acquired and the position of the corpse relative to the robotic system is determined. A medico-legal expert for post-mortem imaging then annotates the desired biopsy targets. For each target, an optimal insertion point is estimated as described in \secref\ref{sec:Robot_Path_Planning}. If no feasible insertion path could be obtained, the CT table is repositioned and the path planning is repeated. In the following, the system setup, the calibration and registration, as well as the tissue extraction are described in detail.

\subsubsection{RPMB System Setup}
The main components that are needed for the operation of the developed planning and decision system are a tracking camera, a robot mounted to a movable table and a medical tomographic imaging system, e.g., CT or MRI. The RPMB system setup is depicted in \figref{\ref{fig:System_Setup_2}, left}. We employ a real-time optical tracking camera (fusionTrack 500, Atracsys LLC, Puidoux, Switzerland) for the tracking of the robot motion and the corpse position. PMCT imaging is performed with a Philips Incisive CT system. The imaging parameters were set to: tube voltage \SI{120}{\kilo\volt}, X-ray tube current \SI{296}{\milli\ampere\second} slice thickness \SI{0.67}{\milli\meter}, spatial resolution along $z$-axis \SI{0.45}{\milli\meter} and pixel spacing \SI{0.98}{\milli\meter}. We use a lightweight medical robot (LBR Med 14, KUKA AG, Augsburg, Germany) which is certified for medical applications and can be used for collaborative operation close to humans. We design and manufacture a custom end effector for mounting biopsy needles which is depicted in \figref{\ref{fig:System_Setup_2}, [A]-[C]}. All parts are CNC manufactured from aluminium blocks. The mounting consists of a bottom plate attached directly to the robot's end effector and a top plate to mount the needle. The plates are connected by a 6-axis force sensor (ATI Industrial Automation, Apex, USA) to record insertion forces. For performing a biopsy we use a Gauge 13 biopsy needle system consisting of a hollow guide needle, a coaxial introducer needle and the biopsy needle for tissue sampling. The guide needle has a length of \SI{190}{\milli\meter}. It provides an effective insertion length of up to \SI{160}{\milli\meter} mounted within the end effector adapter and considering oblique insertions. All needle components are custom manufactured by weLLgo Medical Products GmbH (Wuppertal, Germany). The top plate of the needle mounting has a sliding mechanism with two parts where the top of the biopsy guide needle can be inserted as seen in \figref{\ref{fig:System_Setup_2}, [C]}. 

\begin{figure}[t]
    \begin{tikzpicture}
        \node [] {\includegraphics[width=\linewidth]{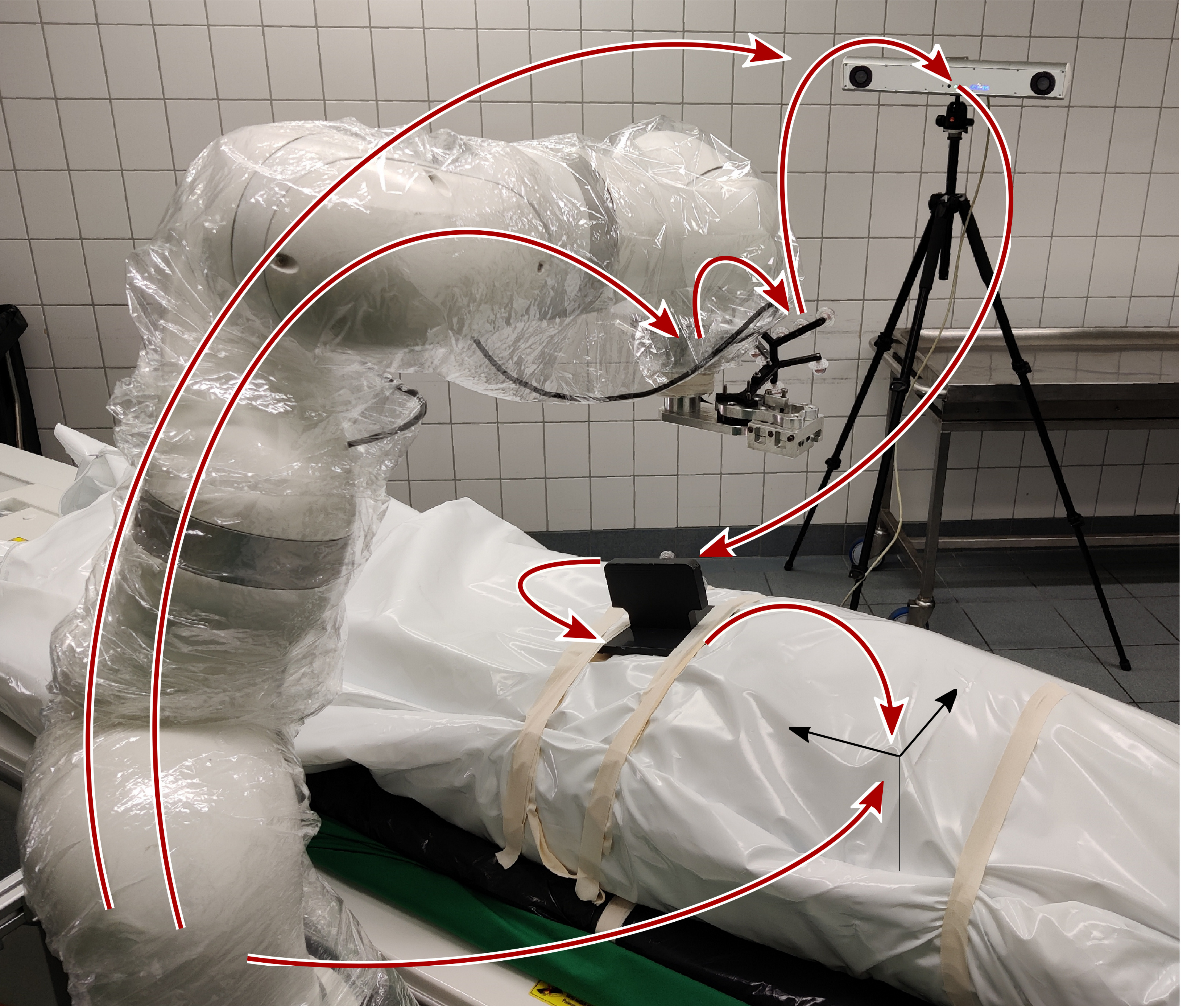}};
        \node[align=left, fill=black!0] at (-3,2.5) {\footnotesize {$_{\text{B}}T^{\text{Cam}}$}};
        \node[align=left, fill=black!0] at (-1.8,1.0) {\footnotesize {$_{\text{B}}T^{\text{EE}}$}};
        \node[align=left, fill=black!0] at (0.6,2.2) {\footnotesize {$_{\text{EE}}T^{\text{M}}$}};
        \node[align=left, fill=black!0] at (2.1,2.5) {\footnotesize {$_{\text{M}}T^{\text{Cam}}$}};
        \node[align=left, fill=black!0] at (3.0,1.2) {\footnotesize {$_{\text{Cam}}T^{\text{R}}$}};
        \node[align=left, fill=black!0] at (-0.5,0) {\footnotesize {$_{\text{R}}T^{\text{SB}}$}};
        \node[align=left, fill=black!0] at (3.0,-1) {\footnotesize {$_{\text{SB}}T^{\text{CT}}$}};
        \node[align=left, fill=black!0] at (0,-2.8) {\footnotesize{$_{\text{B}}T^{\text{CT}}$}};
    \end{tikzpicture}
    \caption{\textbf{Relevant transformations for calibration and registration.} Fixed and variable transformations estimated for calibration and registration. Note the protective body bag which covers the patient to reduce the risk of disease transmission.}
    \label{fig:System_Transformations}
\end{figure}
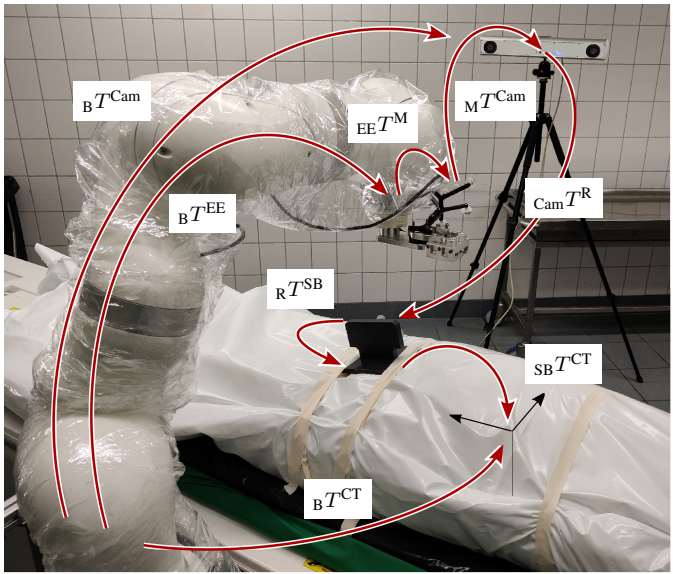

\subsubsection{RPMB System Calibration and CT Registration}
All relevant transformations for calibration of the RPMB system are given in \figref{\ref{fig:System_Transformations}}. We perform a hand-eye calibration to estimate the transformation from the robot base coordinate system (B) to the tracking camera as $_{\text{B}}T^{\text{Cam}}$ and the robot's end effector (EE) to the retro-reflective tracking marker (M) mounted to the custom designed needle holder as $_{\text{EE}}T^{\text{M}}$. Therefore, we record a total of 60 robot poses $_{\text{B}}T^{\text{EE}}$ and tracked poses $_{\text{M}}T^{\text{Cam}}$. To estimate the rigid transforms $_{\text{EE}}T^{\text{M}}$ and $_{\text{B}}T^{\text{Cam}}$, we employ the QR24 algorithm \cite{Ernst.2012}. We then position the corpse on the CT table either in supine or prone position with the head pointing towards the CT gantry. We acquire a PMCT scan covering the desired biopsy targets. For registration of the CT frame relative to the robot's base coordinate system (B), we design a registration phantom which is depicted in \figref{\ref{fig:System_Setup_2}, [D]--[F]}. The phantom is fixed to the corpse with two straps. It consists of a grid of 24 steel balls with radii of \SI{2}{\milli\meter} and \SI{5}{\milli\meter} and four reflective markers for camera tracking. The phantom is CNC drilled into a polyoxymethylene block and the rigid transformation between the steel balls (SB) and the reflective markers (R) is determined from the CAD model as $_{\text{R}}T^{\text{SB}}$. For CT registration, we first segment the steel balls in the CT scan by thresholding and estimating the centroids of the steel balls. Second, we fit the (SB) reference system to the segmented points and estimate $_{\text{CT}}T^{\text{SB}}$. The position of the CT scan relative to the robot's reference system (B) can then be calculated by
\begin{align}
    _{\text{B}}T^{\text{CT}} = {}_{\text{B}}T^{\text{Cam}}\, _{\text{Cam}}T^{\text{R}}\, _{\text{R}}T^{\text{SB}}\, _{\text{SB}}T^{\text{CT}}.
\end{align}
Further, we register the tip of the needle relative to the robot's end effector by driving the needle to the tip of a stylus. The resulting transformation is used as an additional link in the robot setup.

\subsubsection{Tissue Extraction}
In order to minimize the risk of disease transmission during sampling, the corpse is kept inside a protective bag during the entire duration of the procedure. The needles are therefore inserted directly through the plastic and no incision is made into the skin prior to needle entry. Based on the path planning and decision system, the robot moves to the entry position with an offset along the planned trajectory. We secure the guide needle into our custom needle mounting and insert the coaxial introducer needle. This needle features a symmetric tip to reduce lateral bending. To avoid retraction of the introducer needle, we fix a screw onto the top as depicted in \figref{\ref{fig:System_Setup_2}, [B]}. The introducer needle is then inserted into the corpse by the robot to the annotated target location. For safety, the operator needs to confirm each step of the insertion procedure. After the robot has reached its target position, the inner needle is removed by medical staff. The biopsy device is inserted manually until a mechanical stop is reached. The robot is maintaining the position of the guide needle during biopsy extraction to avoid needle movement when puncturing shallow targets. Tissue biopsies with a sample notch of \SI{20}{\milli\meter} are extracted. 

\begin{table*}[t]
    \caption{\textbf{Corpses evaluated in clinical study.} For each corpse gender, body-mass-index (BMI), and if infected with SARS-CoV-2, is given.}
    \label{tab:corpses}
    \centering
    \setlength\tabcolsep{4.5pt} 
    \begin{tabular}{l c c c c c c c c c c c c c c c c c c c c}
        \toprule
        \# & 1 & 2 & 3 & 4 & 5 & 6 & 7 & 8 & 9 & 10 & 11& 12& 13& 14& 15& 16& 17& 18& 19& 20      \\ \midrule
       Sex &  $\male$&	$\male$&	$\male$&	$\male$&	$\male$&	$\male$&	$\male$& $\male$	&$\male$&	$\male$&	$\male$&	$\male$&	$\male$&	$\female$&	$\male$&	$\male$&	$\male$&	$\male$&	$\male$& $\female$
         \\
        Age        & 56 &	81 &	95 &	90 	&79 &	58 &	80& 	78 	&74 &	78&	80 & 52& 	67& 	59 &	83& 58 & 80 & 73 & 70	& 68\\
        BMI    &  25.8  &  25.1 &   14.1  &  22.9 &   20.7 &   20.0 &   24.3 &   30.2 &   26.2 &   35.0  &  24.9  &  37.0  &  25.1 &   31.7 &   19.0 &   23.5 &   18.3 &   28.3  &  23.6 &   29.6    \\
        SARS-CoV-2 & $+$ & $-$ & $+$  & $+$  & $+$  & $-$  & $-$  & $+$  & $-$  & $-$  & $-$  & $-$  & $-$  & $-$  & $-$  & $-$  & $-$  & $-$  & $-$  & $-$ \\
        \bottomrule
    \end{tabular}
\end{table*}

\subsection{Clinical Study for Post-Mortem Sampling}
We evaluate the performance of the system as well as the previously described workflow on 20 corpses. We perform all experiments with medical experts at the Institute of Legal Medicine, University Medical Center Hamburg-Eppendorf, Germany, who are responsible for preparation and handling of the corpses including all biopsy procedures.

\subsubsection{Extended Clinical Workflow for Post-Mortem Sampling}
Additionally to the workflow presented in \secref\ref{sec:clinical_workflow}, we conduct an evaluation of the placed needle positions. For this purpose, we detach the guide needles for each insertion from the end effector mounting while the robot is still in its target position. Once all insertions are made for the current position of the corpse (supine or prone), we acquire a second evaluation PMCT scan. We register the evaluation and the planning PMCT to match the corresponding target annotations. We detect the tip and entry point of each guide needle detected and save their respective 3D coordinates. We consider only a single attempt for the robot-assisted needle insertion for each target. 

\subsubsection{Subjects}
As shown in \tabref\ref{tab:corpses} the corpses evaluated in this study cover a wide range of body types with an average and standard deviation for BMI and age of 25.2$\pm$5.5 and \SI{75(11)}{y}, respectively. 5 out of the 20 subjects in this study were tested positive for SARS-CoV-2. In 8 cases (\SI{40.0}{\%}), we position the subject in the supine position first while we start sampling in the prone position in 12 cases (\SI{60.0}{\%}).

\subsubsection{Needle Placement}
We evaluate the needle placement both with respect to the needle insertion feasibility as well as the needle placement accuracy. For the latter we report the 3D deviation as the distance between the annotated target and the center of the biopsy sample. We define the center of the biopsy to be \SI{10}{\milli\meter} in front of the detected tip along the needle axis of the guide needle. Additionally, we report the lateral deviation as the distance in the direction normal to the needle axis. Results are presented with mean $\pm$ standard deviation. We test significant differences between groups with the Student's t-test or the Wilcoxon rank sum test if the prerequisites for the parametric test are not met. The level of significance is chosen as 0.05. We conduct all analyses in Matlab 2021a (The MathWorks, Inc., Natick, MA, United States).

\subsubsection{Tissue Sampling}
We further report the tissue sampling feasibility via characterization of the obtained samples. We stain the formalin-fixed and paraffine-embedded tissue samples with hematoxylin and eosin. We conduct a histological examination by a forensic pathologist with an optical microscope and report whether the targeted tissue is successfully sampled or if only surrounding tissue is acquired in the attempt. 

\subsubsection{Ethical Approval}
The evaluation study complies with all the relevant national regulations and is in accordance with the tenets of the Helsinki Declaration. The study is approved by the Ethics Committee of the Hamburg Chamber of Physicians and the study protocol includes the informed consent of relatives or legal representatives.

\begin{figure*}[htb]
    \centering
    \includegraphics[width=.9\linewidth,trim=3cm 0 3.0cm 0.9cm,clip]{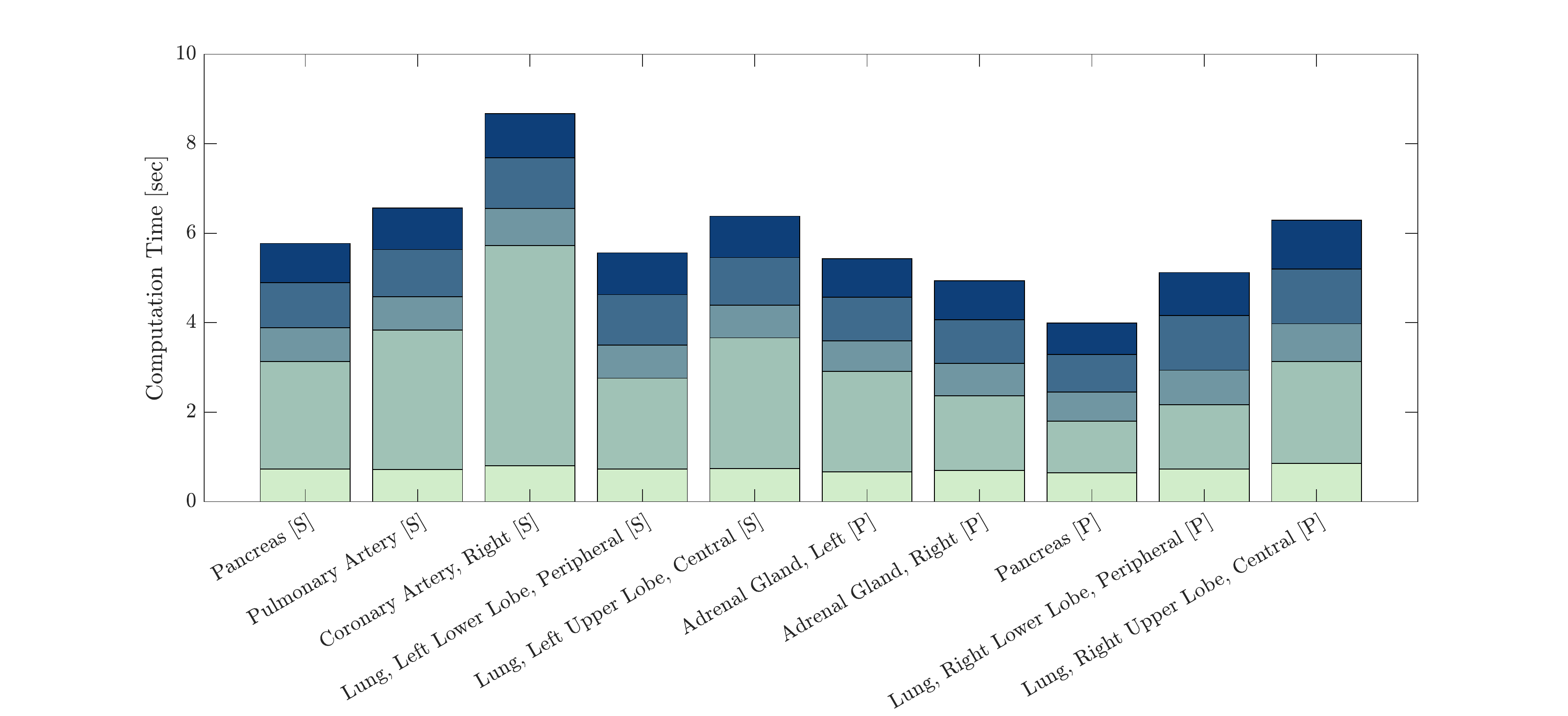}\\
    \caption{\textbf{Mean computation time.} Mean time for each segment of the algorithm for estimating an optimal insertion point. We subdivide possible insertion points in a grid of 30x30mm and check robot reachability for each grid point.}
    \label{fig:runtime_barplot}
\end{figure*}

\section{Results}
In the following, the results of the clinical study will be presented. First, we will present the feasible surface area as well as the mean computation time for our planning and decision system. We will then display the needle placement accuracy, feasibility and biopsy quality for the conducted insertions. For each corpse, we consider 10 targets listed in \tabref{\ref{tab:targets}} for the path planning and subsequent acquisition of biopsy samples.

\begin{table}[t]
\caption{\textbf{Sampled biopsy targets.} The position refers to the corpse placed on the CT table either in supine position (S) or in prone position (P).}
\label{tab:targets}
    \centering
    \begin{tabular}{l c c }
        \toprule
         Biopsy Target & Position & Samples \\
        \midrule
         Pancreas & [S]& 19\\
         Pulmonary Artery & [S] & 18 \\
         Coronary Artery, Right & [S] & 18\\
         Lung, Left Lower Lobe, Peripheral & [S] & 13 \\
         Lung, Left Upper Lobe, Central & [S] & 17 \\
         Adrenal Gland, Right& [P] & 18 \\  
         Adrenal Gland, Left& [P] & 16 \\  
         Pancreasb& [P] & 14\\        
         Lung, Right Upper Lobe, Central& [P]& 18 \\     
         Lung, Right Lower Lobe, Peripheral& [P] & 16 \\
        \bottomrule
    \end{tabular}
\end{table}

\subsection{Planning and Decision System}
We evaluate the computation time on an AMD Ryzen 9 3950X 16-Core processor. We use 32 processes for overlay generation and 30 instances for robot reachability evaluation. Computation times are generally shorter for targets with longer insertion distances because here the decreased number of surface points reduces computational effort more than it gets increased by longer insertion distances per surface point. Computation time for estimating all colormaps is \SI{4.81(156)}{\second} and evaluating robot feasibility for a grid of potential insertion points with a spacing of 30x30mm is \SI{0.91(017)}{\second}.

\begin{figure*}[ht]
    \centering
    \subfloat[Feasible insertion area]{
        \includegraphics[height=80mm]{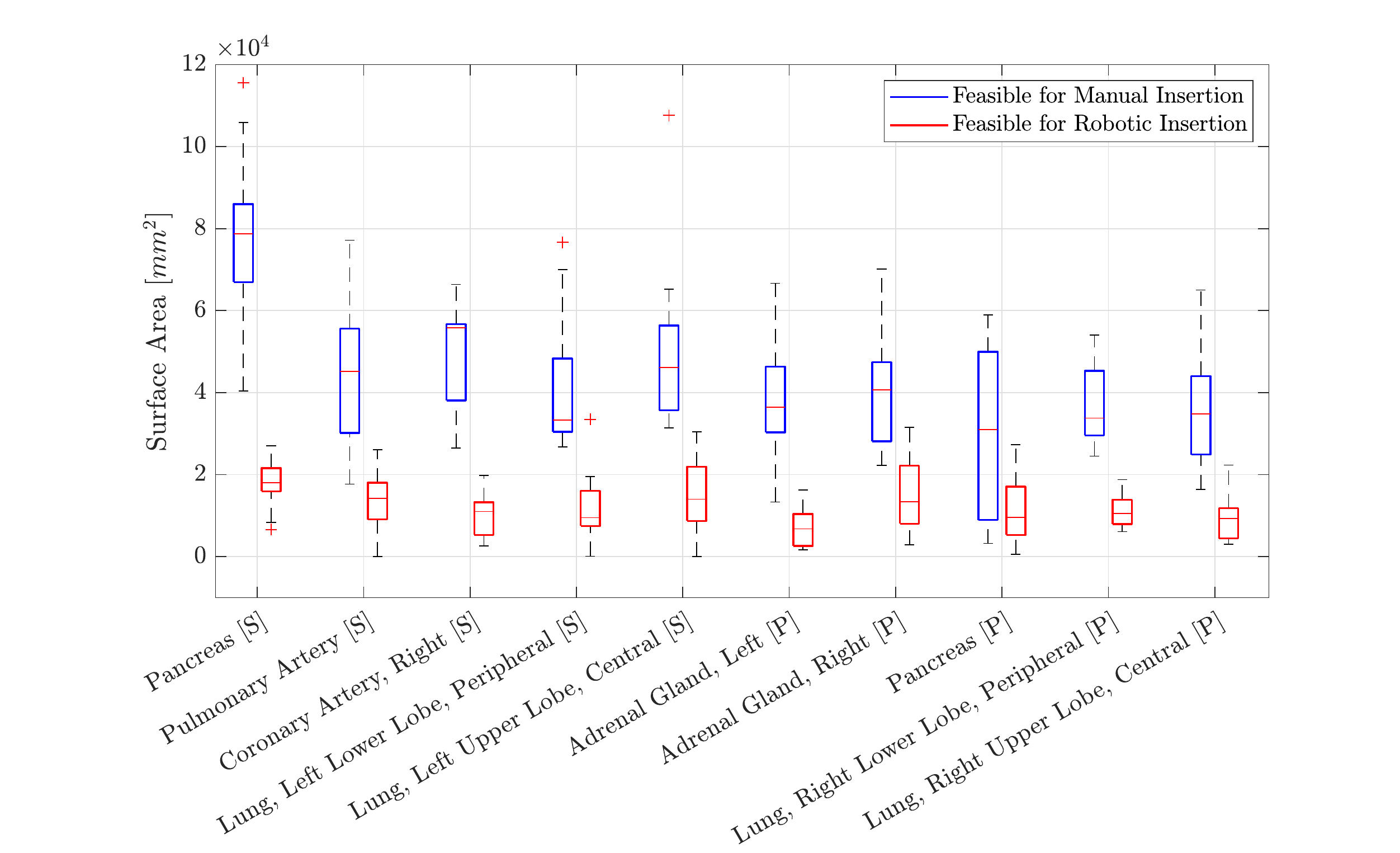}
        \label{fig:Boxplot_AreaRobFeasable}
    }
    \subfloat[Objective value (OV)\\ distribution]{
        \includegraphics[height=80mm]{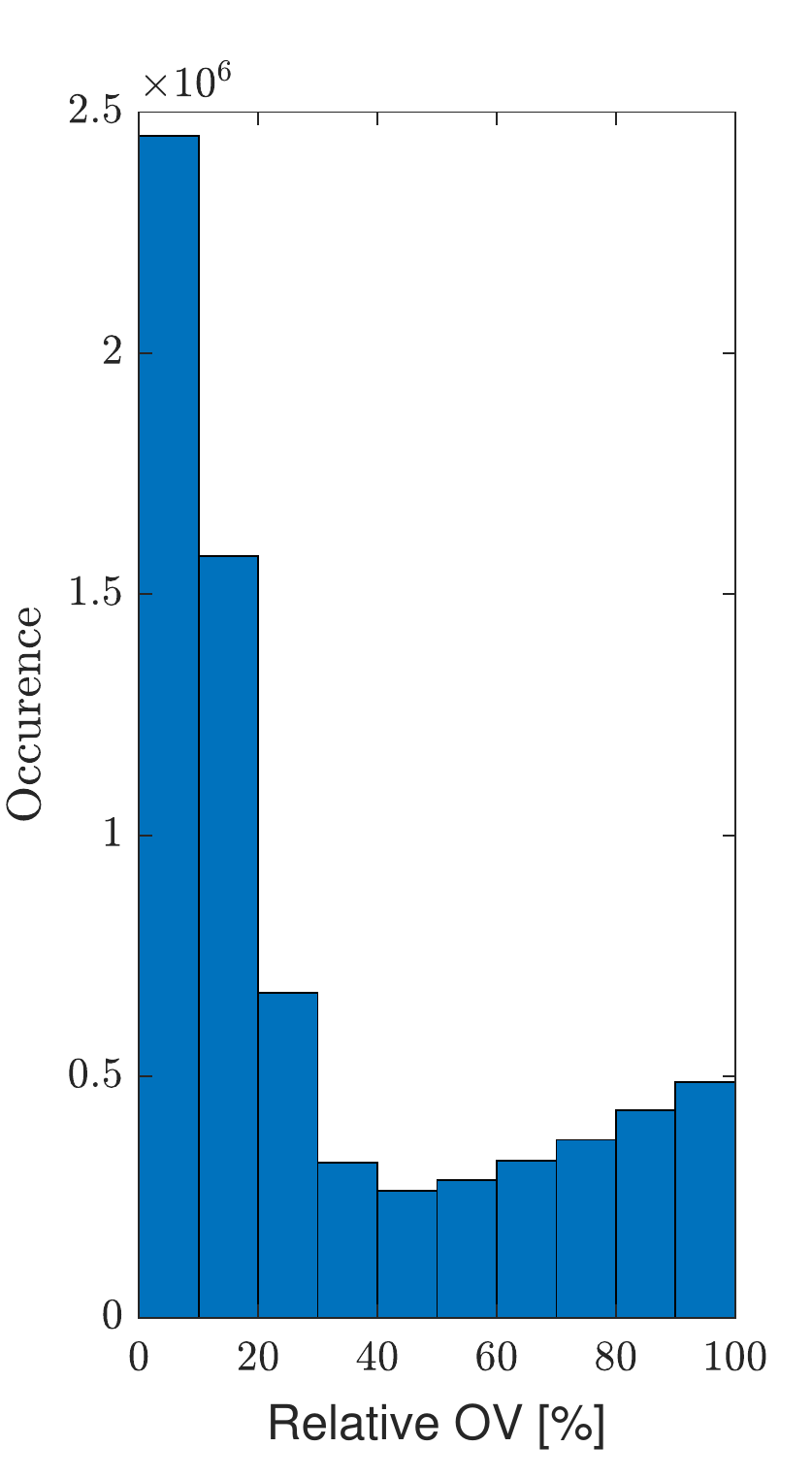}
        \label{fig:Histogram_Occurences}
    }
    \caption{\textbf{Feasible surface area for insertions.} Boxplots for all patients per target are shown and sorted by mean area of manual insertion (a). In blue the surface area without reachability check is shown, i.e the surface area for feasible needle insertion by hand. In red the surface area for feasible robotic needle insertion is shown. The distribution of the objective value at feasible insertion points for all patients and targets is shown on the right (b). Here, the relative objective value, normalized between extrema for the particular patient and target is shown.}
\end{figure*}

From the hand-eye calibration for estimating the robot's base position relative to the tracking camera $_{\text{B}}T^{\text{Cam}}$ we report a mean translation error of \SI{0.77(032)}{\milli\meter} and a mean rotation error of \SI{0.29(014)}{\degree}. For our evaluation of the feasible insertion area, we drive the CT table outside the gantry near the robot base and estimate the transform $_{\text{B}}T^{\text{CT}}$ between robot base and CT. We update the estimate when we move the CT table in case no insertion path is reachable for the robot. In \figref\ref{fig:Boxplot_AreaRobFeasable} the feasible insertion areas for manual and robotic needle insertion for all patients are compared. The feasible area for manual insertion is generally lower for targets with longer insertion distances (pancreas [P], right adrenal gland \ [P]) but can also be small for targets closer to the surface of the skin when insertions have to be made near dense structures (right coronary artery [S], pulmonary artery [S]). The feasible area for robotic insertion mostly follows this trend. However, there are exceptions when targets are hard to reach, e.g. when targets are on the opposite side of the robot (left adrenal gland \ [P], right peripheral lower lung lobe [P]). On average $71.89\%$ of the area feasible by hand is not reachable by the robot. All targets are reachable within a \SI{160}{\milli\meter} insertion distance. We show the distribution of the objective value for feasible insertion points in \figref\ref{fig:Histogram_Occurences}. \SI{34}{\%} of feasible insertion points lie within \SI{10}{\%} of the optimal point.

\subsection{Needle Placement Feasibility}
\begin{figure*}[hbt]
    \setlength\tabcolsep{6pt}
    \centering
    \scriptsize
    \newlength{\biopsyfigheight}
    \setlength{\biopsyfigheight}{0.24\linewidth}
    \begin{tabular}{ccccc}
        \includegraphics[height=\biopsyfigheight]{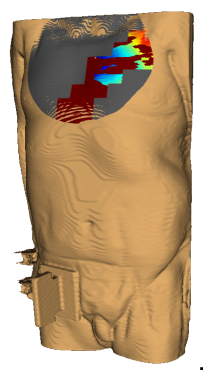} &
        \includegraphics[height=\biopsyfigheight]{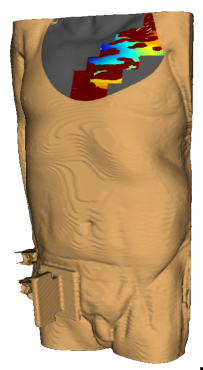} &
        \includegraphics[height=\biopsyfigheight]{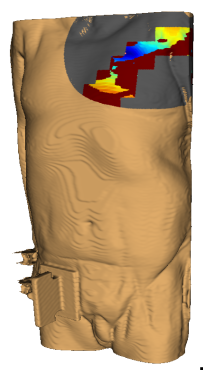} &
        \includegraphics[height=\biopsyfigheight]{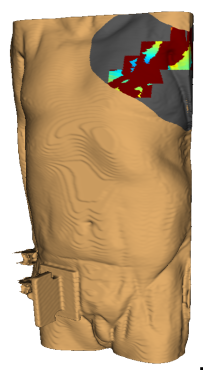} &
        \includegraphics[height=\biopsyfigheight]{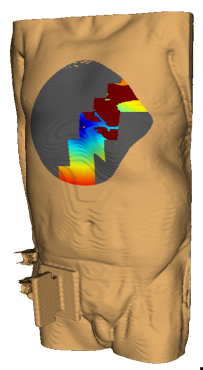} \\
        Right Coronary Artery [S]  & Pulmonary Artery [S] & Lung, L. Up. Lobe, Central  [S] & Lung, L. Lo. Lobe, Peripheral  [S] & Pancreas [S]\\[2mm] 
        \includegraphics[height=\biopsyfigheight]{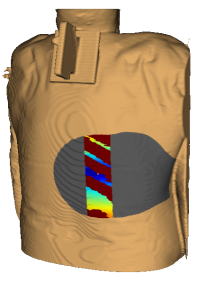} &
        \includegraphics[height=\biopsyfigheight]{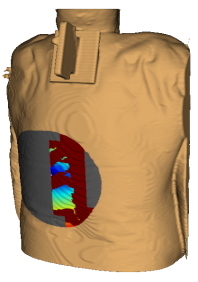} &
        \includegraphics[height=\biopsyfigheight]{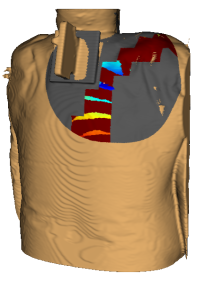} &
        \includegraphics[height=\biopsyfigheight]{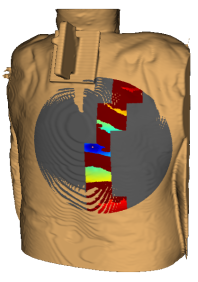} &
        \includegraphics[height=\biopsyfigheight]{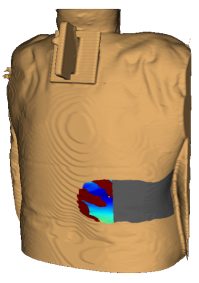}\\
        Right Adrenal Gland [P] &         Left Adrenal Gland [P] &         Lung, R. Up. Lobe, Central [P]  &         Lung, R. Lo. Lobe, Peripheral [P]&         Pancreas [P]\\
    \end{tabular}
    \caption{\textbf{Needle insertion colormaps.} Example of colormaps showing the feasible insertion points and insertion point quality for case 18. Grey points are not reachable by the robot, red points are obstructed by dense structures, blue points represent favourable insertion points.}
    \label{fig:ColormapTargets}
\end{figure*}

\begin{figure}[t]
    \centering
    \includegraphics[width=0.49\linewidth]{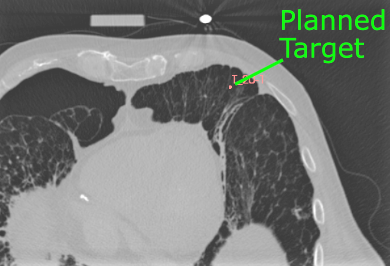} \hfill
    \includegraphics[width=0.49\linewidth]{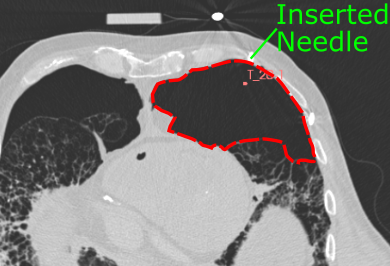}
    \caption{\textbf{Planning vs. Evaluation CT.} Large tissue deformation as a result of a pneumothorax (red dashed line) that causes the anatomy to no longer correlate to the annotated targets.}
    \label{fig:Pneumothorax}
\end{figure}

We consider each of the 10 targets for the biopsy acquisition in all 20 corpses. \figref\ref{fig:ColormapTargets} shows an example of colormaps for all targets of one corpse. Of the 200 planned biopsy targets, we successfully conduct 179 (\SI{89.0}{\%}) robot-controlled insertions. Of the 21 targets without insertions, we do not attempt sampling of lung tissue in four cases (\SI{2.0}{\%}) as a pneumothorax, as seen in \figref{\ref{fig:Pneumothorax}}, is noticed after previous insertions. For two targets (\SI{1}{\%}) we do not attempt the needle insertion due to anatomical obstructions by tumor tissue and bones. For four targets (\SI{2.0}{\%}), we abort the needle insertion because a collision with a previously inserted needle is imminent. For the remaining 11 targets (\SI{5.5}{\%}), insertions are not feasible for other reasons such as limited time with the CT system or failure to find a feasible insertion path for the robot. Out of the 179 insertions, 12 needles are not evaluated in the following analysis because either the needle is moved unintentionally prior to the acquisition of the evaluation PMCT or the scan does not include the entirety of the guide needles. The 167 evaluated insertions into 10 targeted organs are listed in \tabref{\ref{tab:targets}}.

\subsection{Needle Placement Accuracy}
\begin{figure*}[t]
    \centering
    \subfloat[Placement accuracy]{\includegraphics[width=0.87\linewidth,trim=2.5cm 0cm 2.5cm 0cm,clip]{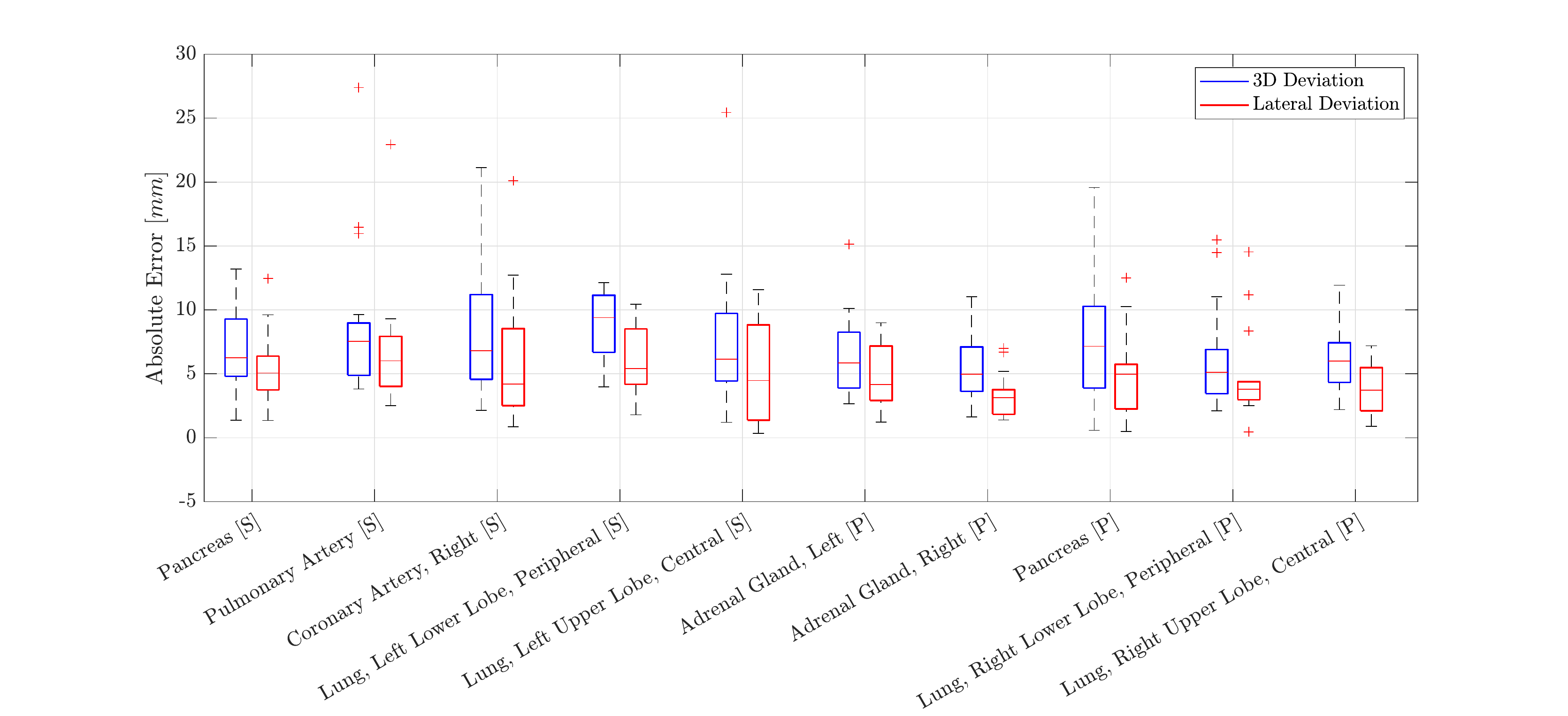}}
    \caption{\textbf{Needle placement accuracy.} Box plots of the absolute error from the annotated target to the center of the sampled biopsy. The 3D deviation is shown in blue and the lateral error normal to the needle axis is displayed in red.}
    \label{fig:Boxplot_AbsoluteAndLateralError}
\end{figure*}

In total, we evaluate 167 robotic needle insertions. The 3D deviation as well as the lateral deviation for each target organ is displayed in \figref\ref{fig:Boxplot_AbsoluteAndLateralError}. Over all attempts, the mean 3D deviation is \SI{7.19(422)}{\milli\meter} and the mean lateral deviation is \SI{5.12(339)}{\milli\meter}. Needles targeting the right adrenal gland in the prone position show the highest accuracy (3D deviation: \SI{5.35(245)}{\milli\meter}, lateral deviation \SI{3.37(169)}{\milli\meter}) while needles targeting the coronary artery result in the lowest accuracy (3D deviation: \SI{9.00(393)}{\milli\meter}, lateral deviation \SI{5.34(288)}{\milli\meter}). We target two organs twice at different sites, once in the supine and once in the prone position. The 3D deviation for needles targeting the pancreas is \SI{7.04(305)}{\milli\meter} and \SI{7.88(519)}{\milli\meter} ($p=.76$) in the supine and prone position respectively. Similarly, for the lower and upper lobe of the lung in the supine and prone position, the 3D deviation is \SI{8.67(286)}{\milli\meter} and \SI{6.26(401)}{\milli\meter} ($p=.04$) as well as \SI{7.78(559)}{\milli\meter} and \SI{6.04(274)}{\milli\meter} ($p=.46$), respectively. We also target the adrenal gland twice, but both in the prone position on the right and on the left side with an accuracy of \SI{5.35(245)}{\milli\meter} and \SI{6.46(329)}{\milli\meter} ($p=.38$), respectively. 

\subsection{Tissue Sampling Feasibility}
\begin{figure*}[h]
    \centering
    \subfloat[Statistical evaluation]{\includegraphics[width = 0.68\linewidth,trim=2.5cm 0cm 3cm 0cm,clip]{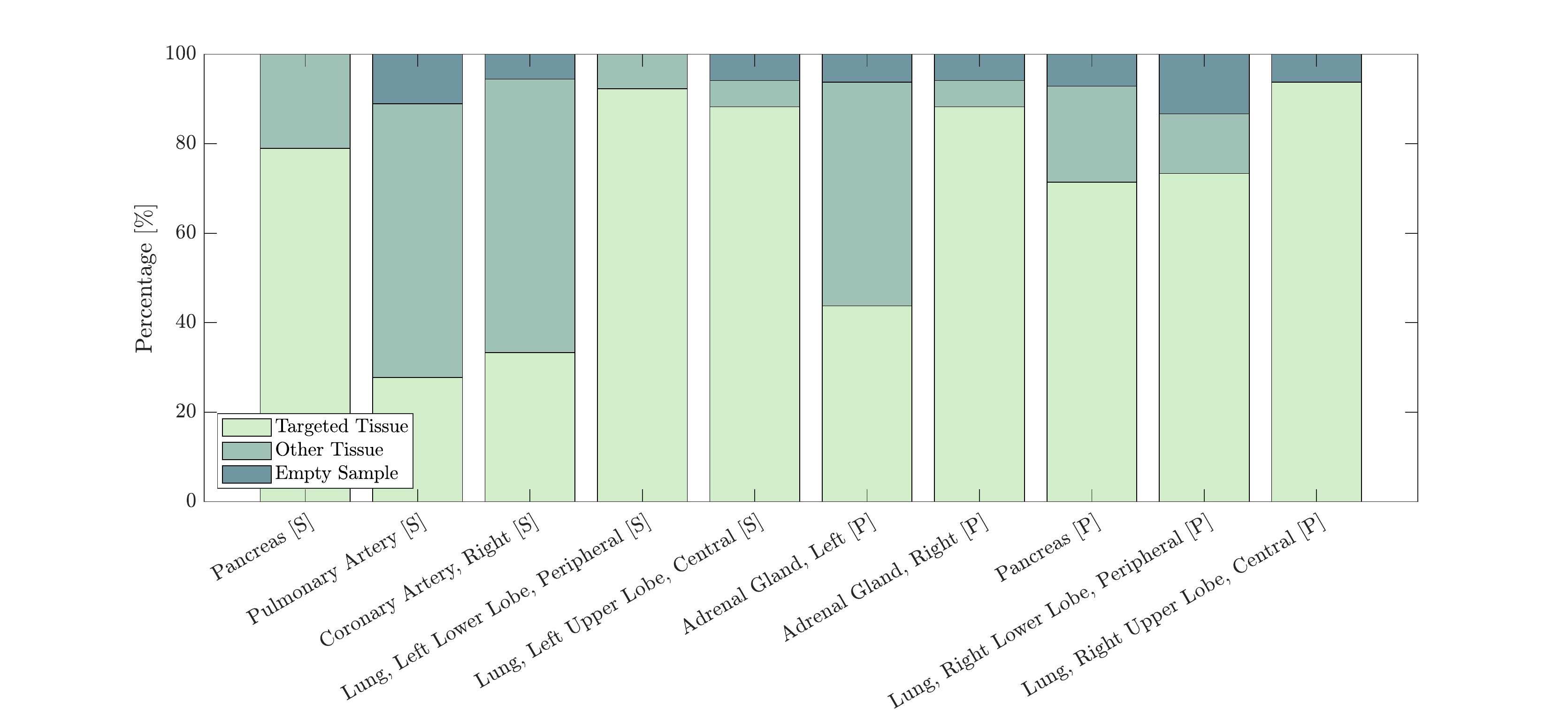}}
    \subfloat[Biopsy samples]{
        \begin{tikzpicture}
            \node [] {\includegraphics[width = 0.3\linewidth]{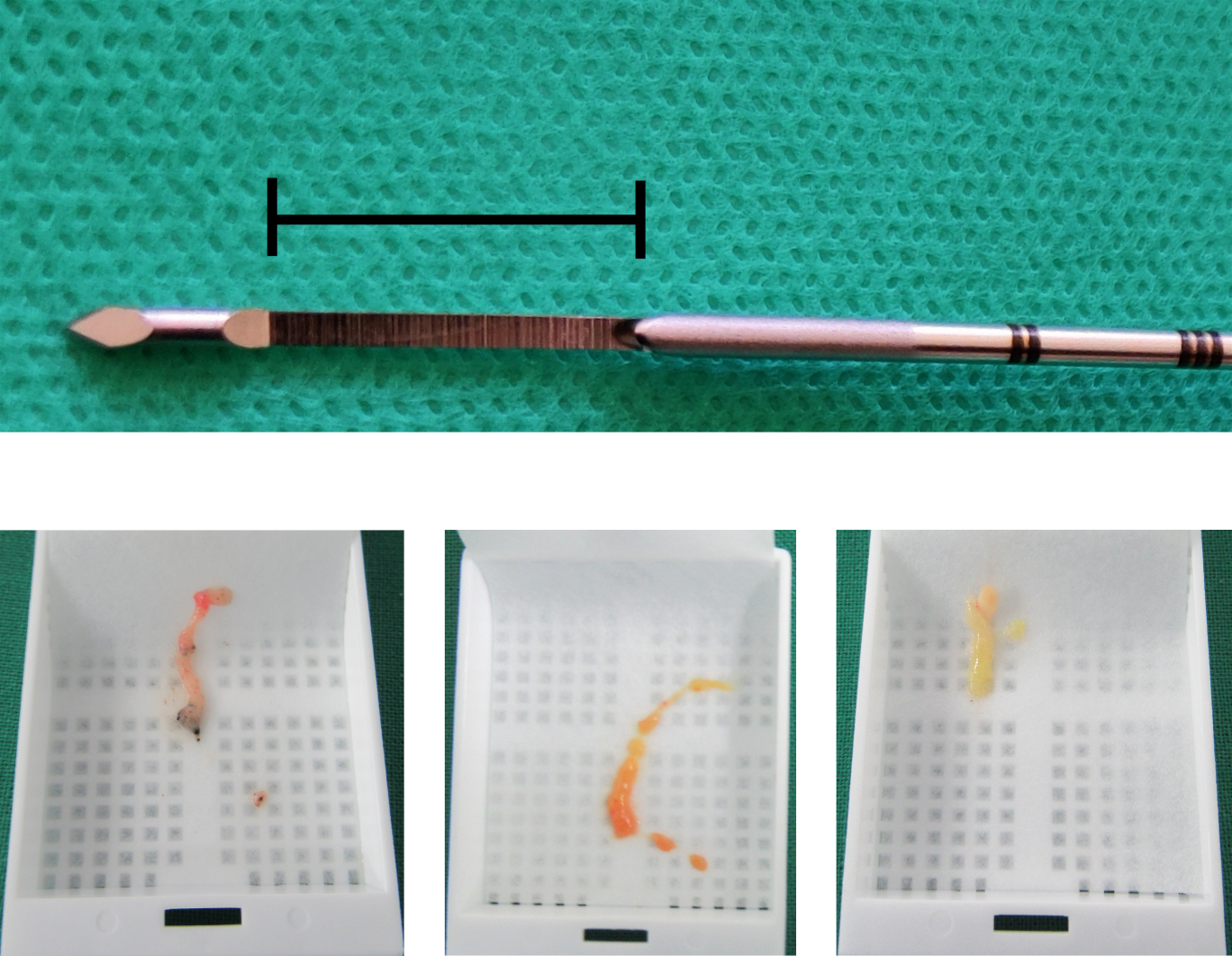}};
            \node [align=left] at (0,-3.66) {}; 
            \node[align=left, rotate=0, fill=white] at (-0.8,1.6) {\SI{20}{\milli\meter}};
            \node[align=left] at (-1.85,2.3) {\footnotesize \textbf{Biopsy Needle}};
            \node[align=left] at (-1.15,-0.06) {\footnotesize \textbf{Extracted Biopsy Samples}};
            \node[align=left] at (-1.9,-2.83) {\footnotesize Lung, Left\\ \footnotesize Upper Lobe, \\\footnotesize Peripheral};
            \node[align=left] at (-0.15,-2.6) {\footnotesize Pancreas \\ };
            \node[align=left] at (1.7,-2.65) {\footnotesize Polmunary\\\footnotesize Artery};
        \end{tikzpicture}
    }
    \caption{\textbf{Histological examination.} For each needle insertion, the acquired sample is characterized and compared to the desired target tissue in a histological examination (a). The employed biopsy sampling needle and biopsy samples are shown in (b).}
    \label{fig:histological_results}
\end{figure*}

Besides the evaluation of the needle placement, we also report the quality of the biopsy samples. The results of the histological examination are displayed in \figref\ref{fig:histological_results}. In \SI{68.1}{\%} of the evaluated cases it is possible to obtain the desired target tissue with a robotic needle insertion in a single attempt. \SI{25.8}{\%} of the acquired samples only contain other tissue from the target's surroundings. \SI{6.1}{\%} of the biopsy samples fail as they render no tissue and cannot be examined. The success rate of the biopsies strongly depends on the chosen target. While we successfully sample the central right upper lobe of the lung in \SI{93.8}{\%} of the cases, the pulmonary artery was sampled only successfully in \SI{27.8}{\%} of cases. The needle placement accuracy for the right adrenal gland is not significantly higher than for the left adrenal gland but it can be seen from \figref\ref{fig:histological_results} that the sample quality is drastically improved given the success rates of \SI{83.3}{\%} and \SI{43.8}{\%}, respectively. For the lung, the cumulative success rate for the lower and upper lobe is similar with \SI{83.95}{\%} and \SI{90.0}{\%} in the prone and supine position, respectively.

\section{Discussion}
We have implemented an open source solution for decision support during needle insertion planning and for execution of robotic biopsy sampling as an alternative to an open autopsy with a 7 DOF robot.

Due to parallelization of heat map generation and feasibility evaluation for robotic insertion, our average runtime is \SI{5.72(167)}{\second} per target. In our evaluation study, the execution time did not prohibit efficient biopsy sampling, especially if heat maps were calculated in the background while the robot still executed the previous needle insertion.  Excluding the feasibility evaluation, our mean \SI{4.81(156)}{\second} runtime is less than the \SI{9.5}{\second} seconds reported by Ebert et al.\ \cite{Ebert.2016} without robot feasibility evaluation. Furthermore, runtime of the system never exceeded \SI{10}{\second} during our evaluation compared to up to \SI{25}{\second} which the system of Ebert et al. could need if insertions were not possible with one orientation. Belbachir et al. report over \SI{40}{\second} planning time \cite{Belbachir.2018}. Reducing the resolution of the CT or grid size of robot feasibility evaluation decreases runtime but may lead to inaccurate heat maps. 

In the current workflow, the CT table was moved by trial and error, if the robot could not reach the desired target, leading to additional delays in the workflow. Therefore, finding a position of the robot requiring as few CT table adjustments as possible is desirable. Additionally, the corpse had to be rotated to reach some annotated targets. However, every interaction of medical staff with the corpse increases the risk of disease transmission. With our custom biopsy needles with a length of \SI{190}{\milli\meter}, we were able to estimate a feasible insertion path for each target evaluated in this study. On average \SI{34}{\%} of feasible insertion points lay within \SI{10}{\%} of the optimal objective value, showing that insertion points with objective values comparable to the optimal point usually exist. However, several targets provided few options for the optimal needle insertion path, limiting needle placement accuracy, potentially prohibiting needle insertion for corpses with higher BMI and if rotation of the corpse should be avoided. While longer needles limit the work space of the robot, which decreases the reachable surface area, the trade-off between longer needles which can offer more feasible insertion area and the option to reach deeper targets but restrict collision-free motion needs to be further studied. In our evaluation study, targets were located in the upper torso. In the future, targets in further body regions as well as narrow trajectories through circumscribed anatomic gaps could be studied.

Currently we do not test for collision with previously inserted needles. Therefore, the operator needs to choose an order of needle insertions which does not lead to collisions. In our evaluation needle collision could not be avoided in \SI{2.0}{\%} of insertions. Note that, our system could be extended for collision detection and insertion order planning similar to Ebert et al. \cite{Ebert.2016}. However, in a real-world clinical workflow, the needles do not need to remain inside the corpse but can be removed after each biopsy. In the future, we would like to to extend our system with an actuated end-effector for fully automated needle insertion, biopsy sampling and needle retraction to further reduce contact time between staff and infectious corpse. Therefore, needle avoidance might not be necessary in a real-world clinical application. Currently, our planning software selects an entry point on the skin automatically based on a cost function. This can be further adapted to also include the uncertainty of a selected insertion point.

The RPMB system is fast to install compared to the system proposed by Ebert et al.~\cite{Ebert.2010} requiring no structural changes to the medical imaging room. Robot and mounting can easily be cleaned or replaced if contaminated. Furthermore, the system can be integrated quickly into the workflow of medical imaging centers due to its fast setup in order to adapt to the dynamics of a pandemic. While the ''VirtoBot'' system can retract to the ceiling to avoid interference with clinical routine medical imaging, the RPMB system can be removed completely from the examination room. The medical expert can use the system remotely to annotate targets while the insertion is performed by the robot and biopsy extraction can be done by medical staff limiting resource usage during a pandemic. Additionally, biopsy targets could also be segmented using convolutional neural networks \cite{Hu.2017} leading to further automation of systematic post-mortem tissue recovery programs/biobanking concepts. Therefore, the presented workflow for our robotic system can be scaled to medical centers at various locations. 

With 20 subjects and 179 completed insertions we thoroughly tested the RPMB system in a real-world scenario. Considering a normal clinical workflow without interfering guide needles and the need for an evaluation CT, even more than the \SI{89.5}{\%} success rate in completed insertions could be feasible. The reported 3D deviation of \SI{7.19(422)}{\milli\meter} and a lateral deviation of \SI{5.12(339)}{\milli\meter} that we report is comparable to similar systems (3D deviation usually ranging from \SI{3}{\milli\meter} to \SI{10}{\milli\meter} \cite{Belbachir.2018, BenDavid.2018, Guiu.2021, Hiraki.2020, Heerink.2019}). However, there are fundamental differences in the setup and biopsy sampling procedure of these systems that influence needle placement accuracy. In comparison to other research groups \cite{Belbachir.2018, Hiraki.2020, Guiu.2021}, we do not make an incision into the skin at the entry point prior to needle insertion. On the contrary, the additional body bag as a safety measure for reducing disease transmissions poses a more challenging needle insertion task. In this study we use a coaxial needle set with a larger diameter compared to needles used in conventional biopsy to compensate for needle bending. This enables the insertion through the body bag without the need of a prior incision with a scalpel. In \cite{BenDavid.2018} and \cite{Hiraki.2020}, needle insertion is conducted under image guidance and adjustments to the trajectory are made during the insertion. However, this limits the operation volume of the robot and requires a custom robot design, being in contrast to the off-the-shelf robot system targeted within this study. Further, we attempt only a single needle insertion and subsequent biopsy sampling for each target and do not allow a second attempt as in \cite{Guiu.2021}. 

In the previously mentioned systems, biopsy sampling is motivated but rarely evaluated. Feasibility of the ``Virtobot'' system has been demonstrated in-vivo in three cases in~\cite{Franckenberg.2021, Staeheli.2016}. To the best of our knowledge, we report verification via histological analysis on a wide range of corpses for the first time for a robotic system. We show, that with the RPMB system it is possible to safely obtain multiple tissue samples from an infected subject. The histological analysis corresponds well with our needle placement evaluation. Vascular tissues, represented by the pulmonary and coronary artery, are more difficult to obtain. This is especially true as organ deformation due to the insertion forces can lead to inaccurate needle placement \cite{Butnariu.2017}. Particularly in thoracic biopsies, relevant gas or fluid volumes can escape through the puncture hole or inserting the needle can result in a gas accumulation in the body cavity (pneumothorax) as seen in \figref\ref{fig:Pneumothorax}. Hence, the planning CT does not match the state of the corpse after insertion and although needles are placed accurately with respect to the planning CT, deformation can lead to incorrect tissue samples being extracted. Therefore, the continuous evaluation of needle forces, detection of tissue displacement during the insertion and the impact of the body bag could improve robotic needle placement accuracy in the future.

\section{Conclusion}
We have conceptualized a system for decision support during needle insertion planning and for execution of robotic biopsy sampling as an alternative to tissue sampling at conventional autopsies. Our results show that the RPMB system facilitates minimally invasive tissue sampling with a robot even under pandemic conditions for casework evaluation in pathology and legal medicine as well as for research-oriented biobanking. Furthermore, we have successfully evaluated the RPMB system in the clinic to acquire biopsy samples from 20 corpses. With an average planning time of \SI{5.72(167)}{\second} per target including evaluation of robot reachability, an average needle placement accuracy of \SI{7.19(422)}{\milli\meter} with a lateral error of \SI{5.12(339)}{\milli\meter} could be achieved. During the clinical evaluation, the desired tissue could be obtained by robotic needle placement in \SI{66.5}{\%} of cases on the first attempt.

\section*{Acknowledgment}
Research funding: This research is partially funded by the \mbox{DEFEAT} \mbox{PANDEMIcs} project from the Bundesministerium für Bildung und Forschung (BMBF) under grant agreement no 01KX2021, partially by the NATON project grant agreement no 01KX2121 and partially by the DFG SCHL 1844/2-2 project. Publishing fees supported by Open Access Publishing of Hamburg University of Technology. \\
Ethical approval: The Ethics Committee of the Hamburg Chamber of Physicians approved the study (No.: 2020-10353-BO-ff).

\ifCLASSOPTIONcaptionsoff
  \newpage
\fi

\bibliographystyle{IEEEtran}
\bibliography{References}

\end{document}